\documentclass[letterpaper, 10 pt, conference]{ieeeconf}  
\IEEEoverridecommandlockouts                              
\usepackage{makecell}
\usepackage{graphics}
\usepackage{epsfig}
\usepackage{color}
\usepackage{amssymb}

\usepackage{array,amsmath,booktabs}
\usepackage{float}
\usepackage{cite}
\usepackage{lipsum}
\usepackage{algorithm}
\usepackage{dblfloatfix}
\usepackage[noend]{algpseudocode}
\usepackage{multicol}
\usepackage{amsmath}
\usepackage{todonotes}
\usepackage{tikz}
\usepackage{mathtools}
\usepackage{array}
\usepackage{makecell}

\DeclareMathOperator{\argmax}{argmax}

\DeclareRobustCommand\sampleline[1]{%
    \tikz\draw[#1, thick] (0,0) (0,\the\dimexpr\fontdimen22\textfont4\relax)
    -- (1em,\the\dimexpr\fontdimen22\textfont4\relax);%
}

\newcolumntype{M}[1]{>{\centering\arraybackslash}m{#1}}

\definecolor{mygreen}{rgb}{0.0,0.4,0.0}
\definecolor{mycyan}{rgb}{0.0,0.72,0.92}
\definecolor{myyellow}{rgb}{0.71,0.65,0.26}
\definecolor{mymagenta}{rgb}{0.54,0.17,0.89}

\graphicspath{ {figures/} }

\title{\LARGE \bf
Decentralized Reinforcement Learning for Multi-Target Search \\and Detection by a Team of Drones*
}

\author{Roi Yehoshua, Juan Heredia-Juesas, Yushu Wu, Christopher Amato, Jose Martinez-Lorenzo$^{1}$
\thanks{*This work is funded by the U.S. Air Force Research Laboratory (AFRL), BAA Number: FA8750-18-S-7007}
\thanks{$^{1}$Khoury College of Computer Science, and Mechanical and Industrial Engineering Department, Northeastern University, Boston, MA 02115, USA \{{\tt\small r.yehoshua, j.herediajuesas, c.amato, j.martinez-lorenzo\}@northeastern.edu, wu.yushu@husky.neu.edu}}%
}

\begin{document}

\maketitle
\thispagestyle{empty}
\pagestyle{empty}

\begin{abstract}

Targets search and detection encompasses a variety of decision problems such as coverage, surveillance, search, observing and pursuit-evasion along with others. In this paper we develop a multi-agent deep reinforcement learning (MADRL) method to coordinate a group of aerial vehicles (drones) for the
purpose of locating a set of static targets in an unknown area. To that end, we have designed a realistic drone simulator that replicates the dynamics and perturbations of a real experiment, including statistical inferences taken from experimental data for its modeling. Our reinforcement learning method, which utilized this simulator for training, was able to find near-optimal policies for the drones. In contrast to other state-of-the-art MADRL methods, our method is fully decentralized during both learning and execution, can handle high-dimensional and continuous observation spaces, and does not require tuning of additional hyperparameters.
\end{abstract}

\section{Introduction}

Recent advancements in unmanned aerial vehicle (UAV) technology have made it possible to use them in place of piloted planes in complex tasks, such as search and rescue operations, map building, deliveries of packages, and environmental monitoring (see \cite{otto2018optimization} for a recent survey).

This paper handles the problem of coordinating a team of autonomous drones searching for multiple ground targets in a large scale environment. The problem of searching and detecting targets in outdoor environments is relevant to many real-world scenarios, e.g., military and first response teams often need to locate lost team members or survivors in disaster scenarios.

Previous methods for target search by UAVs consisted of a division of the surveillance region into cells (e.g., Voronoi cells), and designing a path planning algorithm for each cell \cite{hu2014multi, yang2004decentralized, bertuccelli2005robust}. These methods require direct communication among the drones, often handle poorly online UAV failures, and have no guarantee on the optimality of the final solution. In contrast, we propose a method based on deep reinforcement learning (DRL), which offers an end-to-end solution to the problem. Our method is fully decentralized (does not require any communication between the drones), and guaranteed to converge to a (local) optimum solution. 

While DRL methods have recently been applied to solve challenging single-agent problems \cite{mnih2015human, gu2017deep, hoel2019combining}, learning in multi-agent settings is fundamentally more difficult than the single-agent case due to non-stationarity \cite{hernandez2017survey}, curse of dimensionality \cite{bu2008comprehensive}, and multi-agent credit assignment \cite{agogino2004unifying}.

Despite this complexity, recent multi-agent deep reinforcement learning (MADRL) methods have shown some success, mostly in simple grid-like environments and in game playing \cite{lowe2017multi, foerster2018counterfactual, palmer2018lenient}. Most of existing MADRL methods employ the centralized training with decentralized execution approach, where the agents' policies are allowed to use extra information to ease training, as long as this information is not used at test time. This approach has several limitations, as it assumes noise-free communication between the robots during training, and also it does not allow the agents to adapt their policies to changing environmental conditions during execution (when global information is not available). Moreover, the discrepancy between the information available to the agents during training and execution often leads to instability of the learned policies in runtime.

In this paper we propose a policy gradient MADRL method, which is fully decentralized during both learning and execution. Our method, called Decentralized Advantage Actor-Critic (DA2C), is based on extending the A2C algorithm \cite{mnih2016asynchronous} to the multi-agent case.
To that end, we have developed our own simulator, that is, on one hand, simple and fast enough to generate a large number of sample trajectories; and, on the other hand, realistic enough, accounting for all the dynamics and uncertainties that can affect the deployment of the learned policies on a real team of drones.

We empirically show the success of our method in finding near-optimal solutions to the multi-target search and detection task. To the best of our knowledge, this is the first time that a fully decentralized multi-agent reinforcement learning method has been successfully applied to a large scale, real-world problem.

\section{Related Work}
Hernandez-Leal et al. \cite{hernandez2018multiagent} provide a recent survey of multi-agent deep reinforcement learning (MADRL) methods. They distinguish between value-based methods, that try to learn a state-action value function, and policy gradient methods, that try to optimize the policy function directly without using intermediate value estimates.

Amongst the value-based MADRL methods, two of them are fully decentralized. Decentralized deep recurrent Q-networks (Dec-HDRQN) \cite{omidshafiei2017deep} achieves cooperation by using a smaller learning rate for updates that decrease the Q-value, while Lenient-DQN \cite{palmer2018lenient} achieves cooperation by leniency, optimism in the value function by forgiving suboptimal actions. Both of these methods suffer from sensitivity to hyperparameter values, and can produce only deterministic policies. In contrast, our method generates a fully decentralized stochastic policy, which is useful for handling the exploration/exploitation tradeoff, and does not require any additional hyperparameters to be tuned.

Policy gradient MADRL methods are typically based on the actor-critic architecture, which consists of an actor network that is used to select actions, and a critic network that learns a value function, which is used to update the actor's policy parameters in a direction of performance improvement. All state-of-the-art policy gradient MADRL methods use some form of centralized learning. For example, COMA \cite{foerster2018counterfactual} uses a centralized (shared) critic, MADDPG \cite{lowe2017multi} uses a separate critic for each agent that is augmented with information from other agents, and PS-TRPO \cite{gupta2017cooperative} uses parameter sharing. Contrary to these methods, our method is fully decentralized during both learning and execution, and thus can adapt to changing environmental conditions.

\section{Simulator of People Detection by a Team of Explorer Drones}
\begin{figure}[htp]
	\centering
    \includegraphics[width=0.8\columnwidth,keepaspectratio]{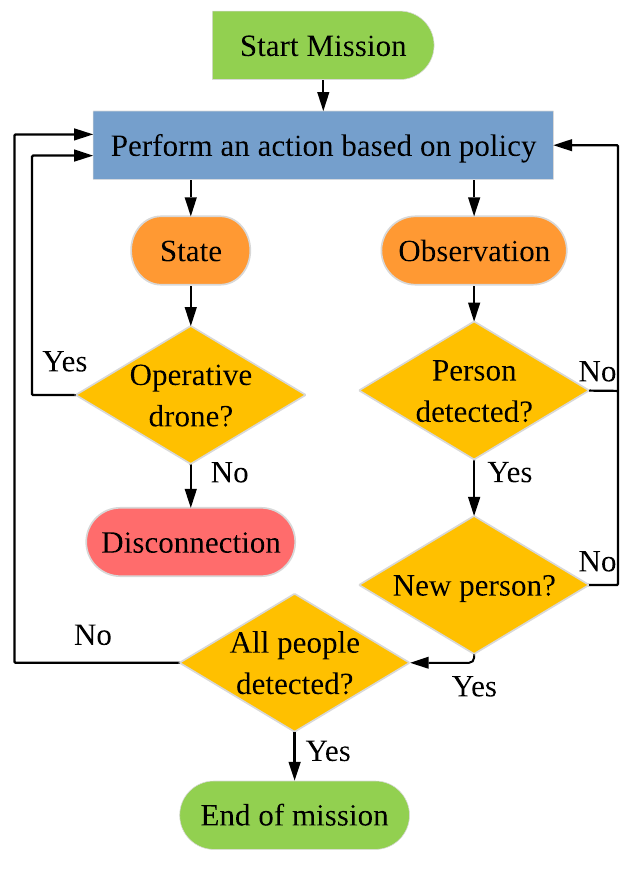}
	\caption{Flow chart for each drone in the team.}
	\label{Flow_chart}
\end{figure}

A 2-D simulator has been designed in order to faithfully replicate the dynamics and detection capabilities of the Intel Aero Ready to Fly Drones. The mission of these drones, working as a team, is to detect and locate the position of a given number of people in a given domain in the most efficient way. In order to successfully accomplish the mission, each drone follows the flow chart described in Fig. \ref{Flow_chart}, which is based on the two main components: states and observations.
These factors determine the actions taken by each drone individually, as well as the global performance of the team.

\subsection{Description of the Domain}
\begin{figure}[tp]
	\centering
    \includegraphics[width=.68\columnwidth,keepaspectratio]{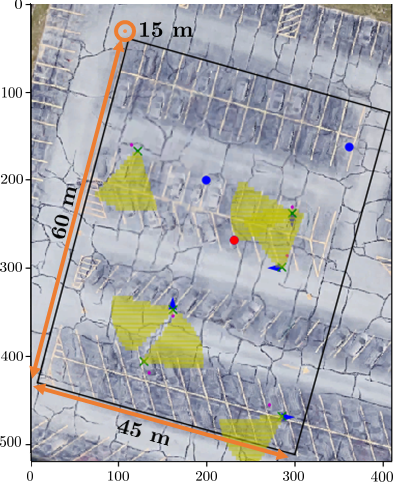}
	\caption{Top view of the drone cage at Kostas Research Institute, where the simulated drones and people are represented.}
	\label{KRI_view}
\end{figure}

The simulator reproduces the drone cage facility located at Kostas Research Institute (KRI), in Burlington, MA. The dimensions of the cage are $60 m\times 45 m \times 15 m$, as shown in Fig. \ref{KRI_view}.
Given that the drones are requested to fly at different but constant altitudes, with enough clearance, 
a 2-D representation of the scene satisfies a realistic approximation, since an overlap in the simulation does not mean a collision. A team of explorer drones equipped with Intel RealSense cameras R200 and a group of people are represented in the scene.

\subsection{Space of States, Observations, and Actions}\label{Space}
\subsubsection{States}
As shown in Fig. \ref{drone_state}, the state of a drone is represented by several elements:

\begin{figure}[tp]
	\centering
	\includegraphics[width=0.7\columnwidth,keepaspectratio]{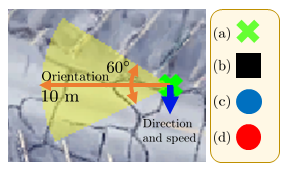}
	\caption{Left: Current state of a drone in the simulator. Right: Legend of symbols, (a) Flying drone, (b) Non-operative drone, (c) Non-detected person, and (d) Detected person.}
	\label{drone_state}
\end{figure}

(i) The shape and color illustrates the mode of flying: a green cross represents a flying drone, meanwhile a black square represents a non-operative drone. 
(ii) A yellow circular sector provides the field of view of the camera of the drone, modeled as explained in section \ref{observations}. Since the camera is located in the front of the drone, this circular sector also indicates its orientation.
(iii) A blue arrow depicts the direction of movement and speed of the drone. Since the drone has the ability of moving in any direction, the orientation and direction do not need to be the same.
(iv) Finally, the drones are equipped with a GPS, so its current position is always known.
The location of the people is represented with blue circles, changing to red when they have been detected by an explorer drone. 

\subsubsection{Observations}\label{observations}
\begin{figure}[tp]
	\centering
	\includegraphics[width=0.95\columnwidth,keepaspectratio]{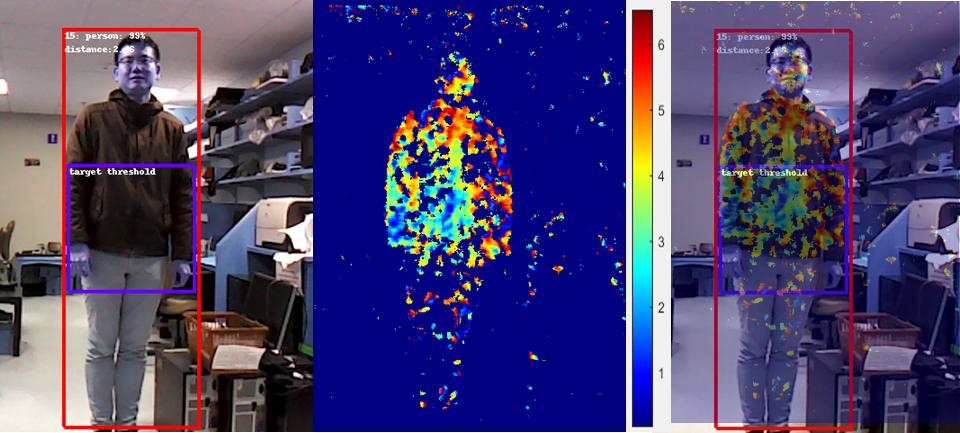}
	\caption{Inference for range estimation. Left: Bounding box of person detected from the RealSense camera. Center: Raw depth information, from $0$ to $6.55m$. (The pixels interpreted farther than the maximum distance are set to 0). Right: Image combination of the raw depth information with the bounding box detection.}
	\label{range_inference}
\end{figure}

The explorer drones perform a continuous observation of the space trying to identify and locate a given number of people in the scene. Each frame collected by the camera is analyzed in real time by the high efficient convolutional neural network (CNN) MobileNets \cite{howard2017mobilenets} to distinguish people among other possible targets, enclosing them into bounding boxes.
The horizontal field of view of the camera, as described in the documentation, is $60^\circ$ \cite{keselman2017intel}, and the range of detection of the camera is estimated to be $10m$, based on field experiments.
The RealSense cameras are also equipped with depth information, which provide the range from the drone to the elements detected on the field of view, as shown in Fig. \ref{range_inference}. In order to determine the distance of the person from the drone, the average of the depth values corresponding to the area of the bounding box, discarding the lower and upper $20\%$ percentiles, is computed.

The combination of the depth information, together with the GPS location of the drone, allows to determine the position of the detected person. The mission is accomplished when the total number of people is detected; but it will fail when all drones crash against the boundaries or when they run out of battery, whose life is estimated to be 15 min (900 s).

\subsubsection{Actions}
There are a total of six basic actions to define the possible behaviour of the drones, organized in two types:

    (i) Direction updates, based on the \textit{NED} commands (North, East, Down). The combination of the \textit{N} and \textit{E} determine the direction of the drone. Since they are set to fly at a constant altitude, the \textit{D} command is kept constant. The four basic actions of this type are the following:
    move North, East, South, and West, all at 1m/s.

    (ii) Orientation updates, based on the \textit{yaw} command. 
    The two basic yaw command actions are     rotate $30^{\circ}$ clockwise and counter clockwise.

Each operating drone is able to perform, at any state, any of these basic actions. 

\subsection{Modeling of uncertainties}
\begin{figure}[tp]
	\centering
	\includegraphics[width=0.6\columnwidth,keepaspectratio]{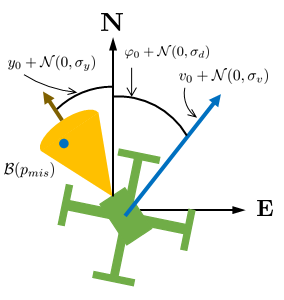}
	\caption{Representation of the uncertainties affecting the flight of the drones.}
	\label{fig:uncertainties}
\end{figure}

A flying drone may be subjected to an enormous amount of uncertainties. In order to perform a realistic simulator, those have to be taken into account. Figure \ref{fig:uncertainties} represents a drone with all the uncertainties considered in the simulator. These uncertainties can be categorized into two main groups: the ones related to the states, and the ones related to the observations.

\subsubsection{State uncertainties}
The position, direction, velocity, and orientation of a drone are subject to external perturbations, such as wind, that disturb their desired values. These perturbations will modify the expected behaviour of the basic actions requested to the drones, in term of the \textit{NED} and \textit{yaw} commands. As explained in Fig. \ref{fig:uncertainties},
the actual values of the direction $\varphi_0$, velocity $v_0$, and yaw $y_0$, will be the results of adding a perturbation to the desired values. These perturbations are modeled by normal distributions with $0$ mean and standard deviations $\sigma_{d},\sigma_{v},\sigma_{y}$, respectively. Since the position of a drone gets determined by its direction and velocity from a previous state, the position uncertainty gets embedded into the ones of the direction and velocity.


\subsubsection{Observation uncertainties}
When a person is in the field of view of the onboard camera, there may be a missdetection, not identifying the person in the scene. This false negative is modeled as a Bernoulli random variable with probability $p_{mis}$.
Since the MobileNets neural network is well trained to identify people, this probability should be very small; however, it will be highly influenced by lighting conditions and partial occlusions. 

\section{Problem Formulation}
\label{PF}
In this section we formalize the multi-target search and detection problem using the Decentralized Partially Observable Markov Decision Process (Dec-POMDP) model \cite{bernstein2002complexity}.

\subsection{Dec-POMDP}
In Dec-POMDP problems, multiple agents operate under uncertainty based on partial views of the world. At each step, every agent chooses an action (in parallel) based on locally observable information, resulting in each agent obtaining an observation and the team obtaining a joint reward.

Formally, the Dec-POMDP model \cite{bernstein2002complexity} is defined by a tuple $\langle  I, S, \{A_i\}, T, R, \{\Omega_i\}, O, h, \gamma \rangle$, where $I$ is a finite set of agents, $S$ is a finite set of states, $A_i$ is a finite set of actions for each agent $i$ with $A=\times_i A_i$ the set of joint actions, $T: S \times A \times S \to [0,1]$ is a state transition probability function, that specifies the probability of transitioning from state $s \in S$ to $s' \in S$ when the actions $\vec{a} \in A$ are taken by the agents, $R: S \times A \to \mathbb{R}^{|I|}$ is an individual reward function, that defines the agents' rewards for being in state $s \in S$ and taking the actions $\vec{a} \in A$, $\Omega_i$ is a finite set of observations for each agent $i$, with $\Omega=\times_i \Omega_i$ the set of joint observations, $O: \Omega \times A \times S \to [0,1]$ is an observation probability function, that specifies the probability of seeing observations $\vec{o} \in \Omega$ given actions $\vec{a} \in A$ were taken which results in state $s' \in S$, $h$ is the number of steps until termination (the horizon), and $\gamma \in [0, 1]$ is the discount factor.

We extended the original Dec-POMDP model by having an individual reward function for each agent, in addition to the global shared reward. This allows the drones to learn the two objectives inherent in the given task: (1) Detect the targets in the shortest time possible, which requires coordination between the drones, and (2) learn to fly within the area boundaries, which is a task that should be learned and thus rewarded by each drone individually. In practice, we combined the shared reward and the individual rewards into a single reward function, that provides the sum of these two rewards for each agent.

A solution to a Dec-POMDP is a \emph{joint policy} $\pi$ --- a set of policies, one for each agent. Because one policy is generated for each agent and these policies depend only on local observations, they operate in a decentralized manner. The value of the joint policy from state $s$ is
\begin{equation}
    V^\pi(s)=\mathbb{E} \left[ \sum_{t=0}^{h-1} \gamma^tR(\vec a^t, s^t) | s,\pi \right],
\end{equation}
which represents the expected discounted sum of rewards for the set of agents, given the policy's actions.

An \emph{optimal policy} beginning at state $s$ is $\pi^*(s) = \argmax_{\pi} V^{\pi}(s)$. That is, the optimal joint policy is the set of local policies for each agent that provides the highest value.

\subsection{Multi-Target Search and Detection}
In this paper, we address the problem of multi-target search and detection by a team of drones. The objective of the drones is to locate and detect the target objects in the minimum time possible, while keeping flying inside the area boundaries. The observations and actions available for each drone are detailed in Section \ref{Space}.

The team gets a high reward (900) for detecting a target, while each drone pays a small cost of -0.1 for every action taken (to encourage efficient exploration), and receives a high penalty (-500) for bumping into the area boundaries.

All the drones start flying from the same region, however, the positions of the targets may change in each episode. In this paper, we assume that there is no explicit communication between the drones, and that they cannot observe each other. Since the positions of the targets are unknown a-priori to the drones, the drones need to find a general strategy for efficiently exploring the environment. Moreover, they need to learn to coordinate their actions, in order not to repeatedly cover areas that have already been explored by other drones.

\section{Approach}
\subsection{Decenetralized Advantage Actor-Critic (DA2C)}
\label{DA2C}
Due to partial observability and local non-stationarity, model-based Dec-POMDP is extremely challenging, and solving for the optimal policy is NEXP-complete \cite{bernstein2002complexity}. Our approach is model-free and decentralized, learning a policy for each agent independently. Specifically, we extend the Advantage Actor-Critic (A2C) algorithm \cite{mnih2016asynchronous} for the multi-agent case. Our proposed method Decentralized Advantage Actor-Critic (DA2C) is presented in Algorithms \ref{alg:DA2C} and \ref{alg:TrainAgent}.

A2C is a policy gradient method, that targets at modeling and optimizing the policy directly. The policy is modeled with a parameterized function with respect to $\theta$, $\pi_\theta(a|s)$. The objective value of the reward function depends on this policy, and can be defined as: $J(\theta) = \sum_{s \in S} d^\pi(s)V^\pi(s)$, where $d^\pi(s)$ is the stationary distribution of states.

According to the policy gradient theorem \cite{sutton2018reinforcement},
\begin{equation}
    \nabla_\theta J(\theta) = \mathbb{E}_{s, a \sim \pi}[Q^\pi(s,a) \nabla_\theta \log \pi_\theta(a|s)]
\end{equation}

A main limitation of policy gradient methods is that they can have high variance \cite{konda2000actor}. The standard way to reduce the variance of the gradient estimates is to use a baseline function $b(s)$ inside the expectation:

\begin{equation}
    \nabla_\theta J(\theta) = \mathbb{E}_{s, a \sim \pi}[(Q^\pi(s,a) - b(s)) \nabla_\theta \log \pi_\theta(a|s)]
\end{equation}

A natural choice for the baseline is a learned state-value function $b(s) = V^\pi(s)$, which reduces the variance without introducing bias. When an approximate value function is used as the baseline, the quantity $A(s, a) = Q(s, a) - V(s)$ is called the \emph{advantage function}. The advantage function indicates the relative quality of an action compared to other available actions computed from the baseline.

In actor-critic methods \cite{konda2000actor}, the actor represents the policy, i.e., action-selection mechanism, whereas a critic is used for the value function learning. The critic follows the standard temporal difference (TD) learning \cite{sutton2018reinforcement}, and the actor is updated following the gradient of the policy's performance.

Thus, the loss function for A2C is composed of two terms: policy loss (actor), $\mathcal{L}_\pi$, and value loss (critic), $\mathcal{L}_v$. An entropy loss for the policy, $H(\pi)$, is also commonly added, which helps to improve exploration by discouraging premature convergence to suboptimal deterministic policies. Thus, the loss function is given by:
\begin{equation}
    \mathcal{L} = \lambda_\pi \mathcal{L}_\pi + \lambda_v \mathcal{L}_v - \lambda_H \mathbb{E}_{s \sim \pi}[H(\pi(\cdot|s))]
\end{equation}

with $\lambda_\pi, \lambda_v, \lambda_H$ being weighting terms on the individual loss components.

\begin{figure}
  \centering
  \includegraphics[width=1\columnwidth,keepaspectratio]{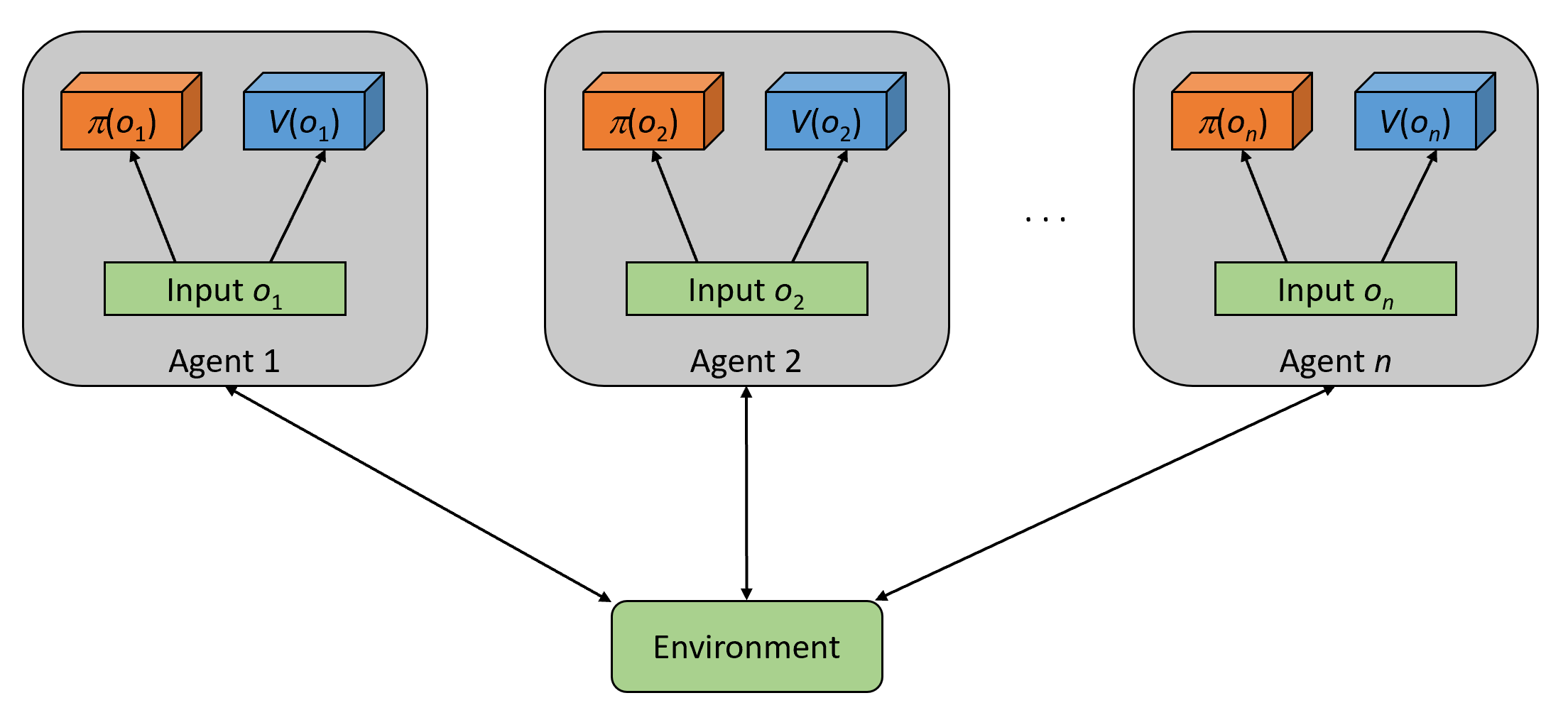}
  \caption{Overview of our multi-agent decentralized actor, critic approach.}
  \label{fig:architecture}
\end{figure}

The architecture of our decentralized actor-critic algorithm is depicted in Figure \ref{fig:architecture}. As described in Algorithm \ref{alg:DA2C}, our training process alternates between sampling trajectories by the team of agents (lines 7--14), and optimizing the networks of the agents with the sampled data (lines 17--23). In the procedure \textsc{TrainAgent} described in Algorithm \ref{alg:TrainAgent}, we accumulate gradients over the mini-batch of samples, and then use them to update the actor and critic networks' parameters. Accumulating updates over several steps provides some ability to trade off computational efficiency for data efficiency.

\begin{algorithm}[h]
    \small
    \caption{DA2C ($I$, $T_{max}$, $m$)}
    \label{alg:DA2C}
    \begin{algorithmic}[1]
        \Require A group of agents $I$, maximum episode length $T_{max}$, and batch size $m$

        \For{$i \in I$}
            \State Initialize actor and critic network weights $\theta_i$, $\omega_i$
        \EndFor
        \State Get an initial joint observation $\vec{o}$

        \Repeat
            \State $t \gets 1$
            \State Initialize buffer $\mathcal{B}$
            \State // Collect $m$ samples from the environment

            \For{$j \gets 1,...,m$}
                \For{$i \in I$}
                    \State Sample action $a_i \sim \pi_{\theta_i}(\cdot|o_i)$
                \EndFor

                \State Execute the joint action $\vec{a} = (a_1, ..., a_n)$
                \State Receive a new joint observation $\vec{o'}$ and reward $\vec{r}$
                \State Store transition ($\vec{o}, \vec{a}, \vec{r}, \vec{o'}$) in $\mathcal{B}$
                \State $t \gets t + 1$
            \EndFor

            \State // Train the agents using the samples in the buffer

            \For{$i \in I$}
                \State Initialize $O_i, A_i, R_i, O'_i$ to empty sets
                \For{each transition $(\vec{o}, \vec{a}, \vec{r}, \vec{o'}) \in \mathcal{B}$}
                    \State $O_i \gets O_i \cup \{\vec{o}_i\}$
                    \State $A_i \gets A_i \cup \{\vec{a}_i\}$
                    \State $R_i \gets R_i \cup \{\vec{r}_i\}$
                    \State $O'_i \gets O'_i \cup \{\vec{o'}_i\}$
                \EndFor

                \State \textsc{TrainAgent}($O_i, A_i, R_i, O'_i$)
            \EndFor

            \State $\vec{o} \gets \vec{o'}$
        \Until $t > T_{max}$ \text{or mission accomplished}
    \end{algorithmic}
\end{algorithm}

\begin{algorithm}[h]
    \small
    \caption{\textsc{TrainAgent}($O, A, R, O', i, m$)}
    \label{alg:TrainAgent}
    \begin{algorithmic}[1]
        \Require A sequence of observations $O$, actions $A$, rewards $R$, and next observations $O'$,
             agent index $i$, and batch size $m$
        \State // Initialize the variable that holds the return estimation
        \vspace*{0.2em}
        \State $G \gets \begin{cases}
                             0 & \text{if $s_m$ is a terminal state} \\
                             V_{\omega_i}(o'_m) & \text{otherwise}
                        \end{cases}$

        \For{$j \gets m - 1,...,1$}
            \State $G \gets \gamma G + r_j$
            \State Accumulate gradients w.r.t. $\theta_i$: \newline
                \hspace*{2em} $d{\theta_i} \gets d{\theta_i} + \nabla_{\theta_i} \log \pi_{\theta_i}(a_j|o_j)(G - V_{\omega_i}(o_j))$
            \State Accumulate gradients w.r.t. $\omega_i$: \newline
                \hspace*{2em} $d{\omega_i} \gets d{\omega_i} + 2(G - V_{\omega_i}(o_j)) \nabla_{\omega_i}(G - V_{\omega_i}(o_j))$
        \EndFor
        \State Update $\theta_i$ using $d{\theta_i}$, and $\omega_i$ using $d{\omega_i}$
    \end{algorithmic}
\end{algorithm}

\subsection{Network Architecture}
Each drone has two neural networks: one for the actor and one for the critic. Both networks consist of three fully connected layers with ReLU nonlinearities. The first layer has 200 neurons and the second one has 100 neurons. The output of the actor network is a probability distribution over the actions, thus its output layer has six neurons (one for each possible action), whereas the critic network returns a single number, which represents the approximate state value.

\section{Results}
Our first experiment involves an environment with three drones and three targets, where all the drones start flying from the bottom left corner of the area. The parameters of Algorithm \ref{alg:DA2C} and the training process are shown in Table \ref{tab:parameters}.

\begin{small}
\begin{table}[h]
    \centering
    \caption {Parameters of DA2C}
    \begin{tabular}{ l l r }
        \hline
        Discount factor & $\gamma$ & 0.99 \\
        Learning rate & $\eta$ & 0.0001 \\
        Mini-batch size & $m$ & 32 \\
        Policy loss weight & $\lambda_\pi$ & 1 \\
        Value loss weight & $\lambda_v$ & 1 \\
        Entropy loss weight & $\lambda_H$ & 0.001 \\
        Maximum episode length & $T_{max}$ & 900 \\
        Drone's direction std & $\sigma_{d}$ & 0.1 \\
        Drone's orientation std & $\sigma_{y}$ & 0.1 \\
        Drone's speed std & $\sigma_{v}$ & 0.1 \\
        Misdetection probability & $p_{mis}$ & 0.05 \\
        \hline
    \end{tabular}
    \vspace{-1.0em}
    \label{tab:parameters}
\end{table}
\end{small}

\begin{figure}
  \centering
  \includegraphics[width=0.8\columnwidth,keepaspectratio]{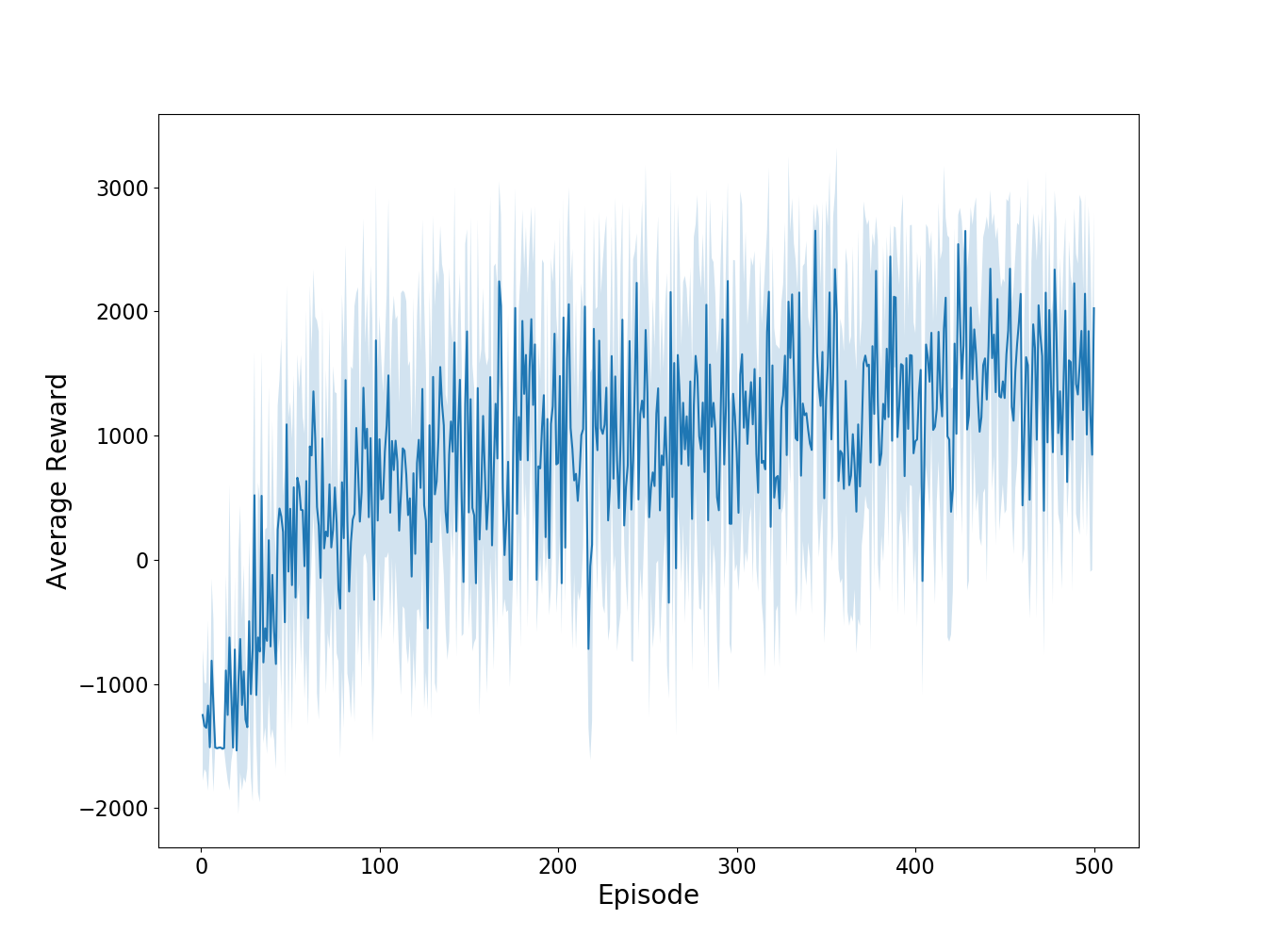}
  \caption{Average reward and standard deviation per episode in an environment with three drones and three targets.}
  \label{fig:da2c_3drones}
\end{figure}

Figure \ref{fig:da2c_3drones} shows the average reward $\bar{r}$ and standard deviation per episode for 500 training episodes. The average is computed over five independent runs with different random seeds. Each training session took approximately 5 hours to complete on a single Nvidia GPU GeForce GTX 1060.

The maximum possible reward that can be attained in this scenario is $900 \cdot 3 - (0.1 \cdot 3) n = 2700 - 0.3n$, where $n$ is the number of time steps it takes for the drones to detect all the targets. Since the maximum length of an episode is 900 time steps, the maximum possible reward lies in the range [2430, 2700], depending on the initial locations of the targets. As can be seen in the graph, after a relatively small number of episodes (about 400 episodes), the team was able to reach an average reward very close to the maximum (2648). The fluctuations in the graph can be attributed to the fact that some of the initial configurations of the targets are significantly harder to solve than others (e.g., when the targets are located in different corners of the environment).

By examining the learned policies of the drones, we can see that the work area is first split between the drones, and then each drone thoroughly explores its own subarea by simultaneously moving and rotating the camera for maximum coverage efficiency. A sample video showing the final joint policy is provided in the supplemental material.

Next, we compared the performance of our learned joint policy against two baselines. In the first baseline, the drones choose their actions completely randomly. The second baseline is a collision-free policy, where the drones fly randomly most of the time, but change their direction by 180 degrees when they get near the walls. Note that this baseline has an edge over our learned policy, as our drones had to learn not to collide with the walls.

\begin{figure}
  \centering
  \includegraphics[width=0.9\columnwidth,keepaspectratio]{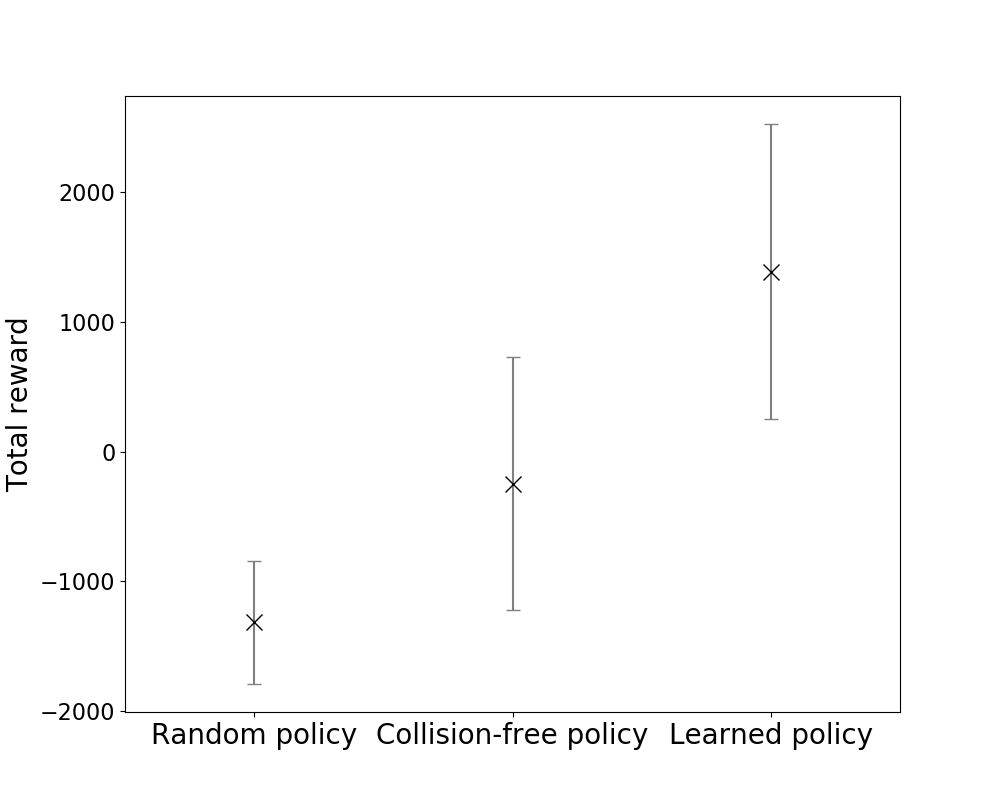}
  \caption{The total reward and standard deviation achieved by our learned policy vs. a random policy and a collision-free policy, averaged over 500 episodes.}
  \label{fig:random_vs_learned}
\end{figure}

All three policies (the learned one and the two baselines) have been evaluated on 500 episodes with different initial locations of the targets. Figure \ref{fig:random_vs_learned} shows the results. As can be seen, our learned policy significantly outperforms the two baselines, achieving a mean total reward of 1388.36, while the total mean reward achieved by the random policy and the collision-free policy are -1314.72 and -247.56, respectively.

\begin{figure}
  \centering
  \includegraphics[width=0.8\columnwidth,keepaspectratio]{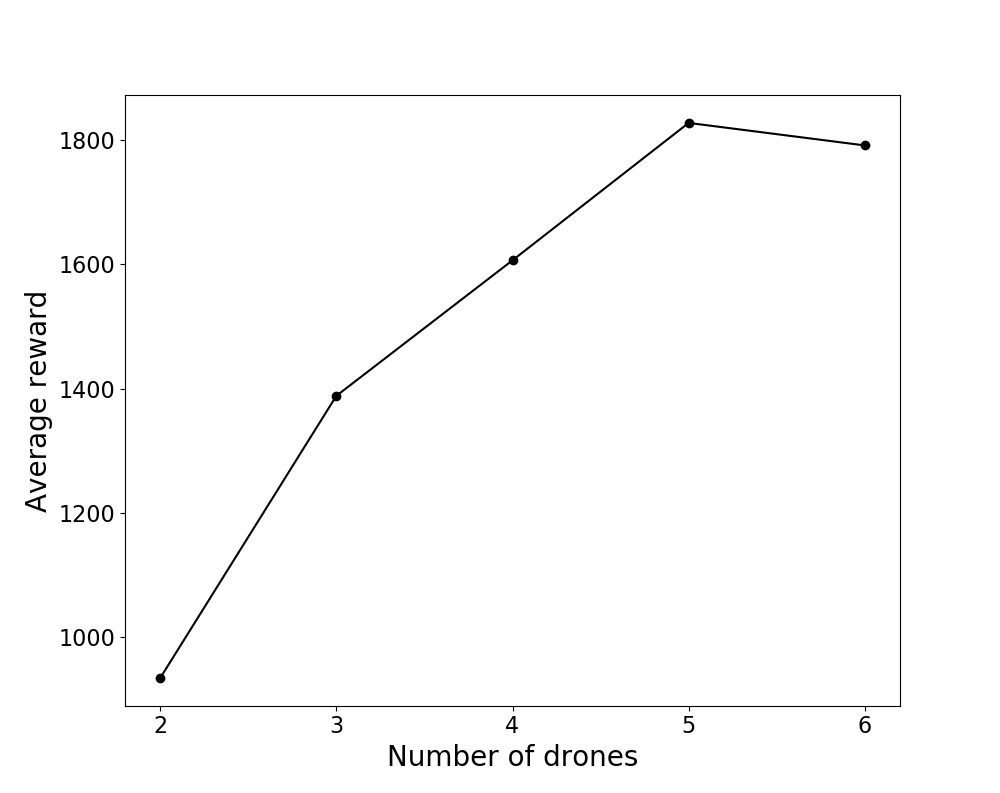}
  \vspace{-1.0em}
  \caption{The average reward achieved by different team sizes, ranging from two drones to six drones.}
  \label{fig:changing_team_size}
\end{figure}

We have also have examined the impact of changing the number of drones in the team on the team's ability to fulfill the task. Figure \ref{fig:changing_team_size} shows the average reward achieved by different team sizes, ranging from two drones to six drones. The number of targets remained three in all experiments. Clearly, adding more drones to the team increases the probability of detecting all targets within the time limit. However, increasing the team size for more than five drones does not improve the performance any further, which implies that the team has reached a near-optimal solution (a team with five drones was able to achieve an average reward of 1827 over 500 evaluation runs).

\begin{figure}
  \centering
  \includegraphics[width=0.8\columnwidth,keepaspectratio]{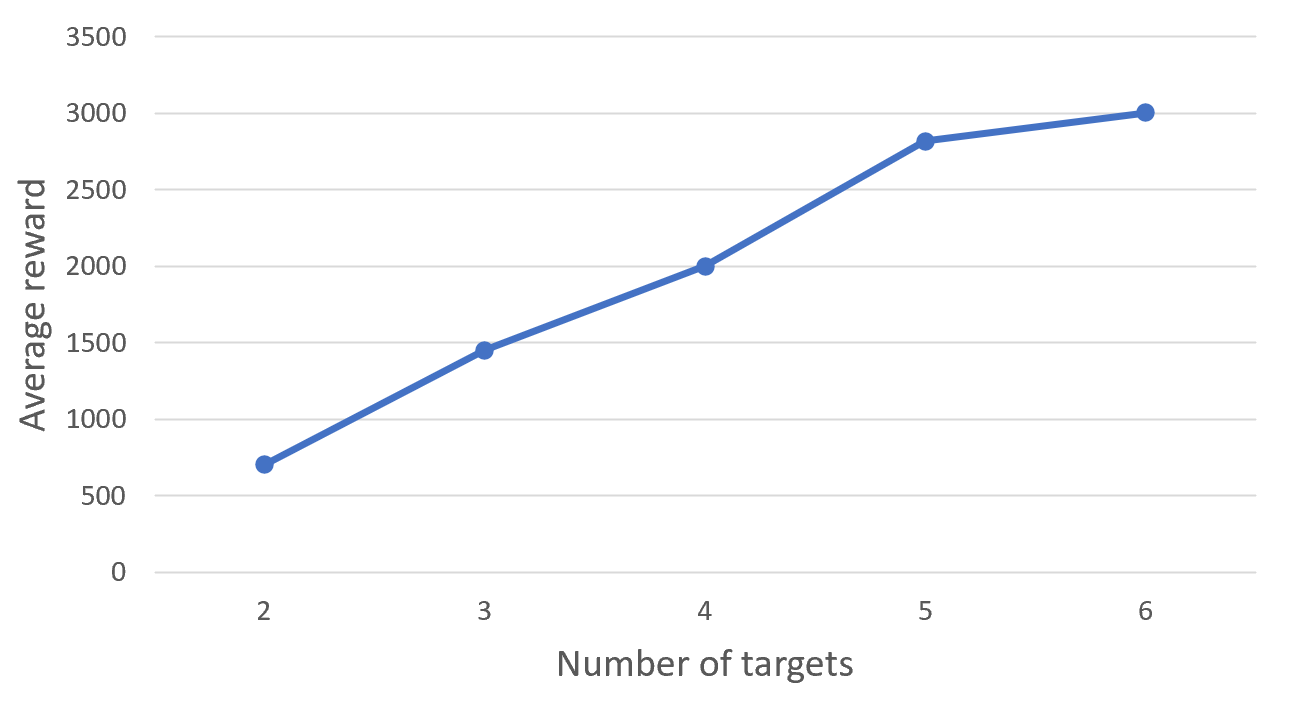}
  \caption{The average reward achieved by a team of 3 drones for various number of targets.}
  \label{fig:changing_targets_num}
\end{figure}

Lastly, we have examined the ability of the drones to detect different numbers of targets. Figure \ref{fig:changing_targets_num} shows the average reward achieved by a team of three drones, trying to detect between two to six targets. We can observe an almost linear relationship between the number of targets and the average return, which means that the time required to find any additional target is nearly constant.

\section {Conclusion}
We have proposed a fully decentralized multi-agent policy gradient algorithm to solve a challenging real-world problem of multi-target search and detection. Our method is able to find a near-optimal solution to the problem using a short training time. Despite being completely decentralized, our drones learn to coordinate their actions as to minimize the overlap between the areas they are exploring.

In the future, we would like to consider dynamic environments, in which the targets may change their locations during execution, as well as implement more complex models that account for external uncertainties such as wind or illumination. We also intend to add more sensors to the drones, extend the simulator to 3D, and test the results on real drones.





\bibliographystyle{IEEEtran}
\bibliography{drones}

\end{document}